\documentclass[twoside,11pt]{article}
\usepackage[abbrvbib]{dmlr2e}
\usepackage{amsmath}
\usepackage{xcolor}
\usepackage{hyperref}
\usepackage{cleveref}
\usepackage{booktabs}
\usepackage{graphicx}
\usepackage{float}
\usepackage{caption}
\usepackage{subcaption} 
\usepackage{hyperref}       % hyperlinks
\hypersetup{
    colorlinks=true, % 激活链接颜色
    linkcolor=blue,  % 文档内部链接颜色（例如目录、\ref和\pageref）
    filecolor=magenta, % 文件链接颜色
    urlcolor=cyan,   % 外部URL链接的颜色
    citecolor=cyan   % 引用的颜色（例如文献引用）
}

\usepackage{lastpage}
\def\openreview{\url{https://openreview.net/forum?id=n3R1texieu}} % Insert correct link to OpenReview for camera-ready version
\ShortHeadings{FAIntbench}{Luo et al.}
\firstpageno{1}
\begin{document}
\title{FAIntbench: A Holistic and Precise Benchmark \\for Bias Evaluation in Text-to-Image Models}

\author{\name Hanjun Luo \email hanjun.21@intl.zju.edu.cn \\
        \name Ziye Deng \email ziye.21@intl.zju.edu.cn \\
        \name Ruizhe Chen \email ruizhec.21@intl.zju.edu.cn \\
        \name Zuozhu Liu \thanks{Corresponding author} \email zuozhuliu@intl.zju.edu.cn \\
       \addr Zhejiang University\\
       Hangzhou, Zhejiang, China
        }
\editor{ICML 2024 Workshop DMLR Reviewers}
\maketitle

\begin{abstract}
The rapid development and reduced barriers to entry for Text-to-Image (T2I) models have raised concerns about the biases in their outputs, but existing research lacks a holistic definition and evaluation framework of biases, limiting the enhancement of debiasing techniques. To address this issue, we introduce FAIntbench, a holistic and precise benchmark for biases in T2I models. In contrast to existing benchmarks that evaluate bias in limited aspects, FAIntbench evaluate biases from four dimensions: manifestation of bias, visibility of bias, acquired attributes, and protected attributes. We applied FAIntbench to evaluate seven recent large-scale T2I models and conducted human evaluation, whose results demonstrated the effectiveness of FAIntbench in identifying various biases. Our study also revealed new research questions about biases, including the side-effect of distillation. The findings presented here are preliminary, highlighting the potential of FAIntbench to advance future research aimed at mitigating the biases in T2I models. Our benchmark is publicly available to ensure the reproducibility.\\
\end{abstract}

\section{Introduction}
As one of the crucial multi-modal technologies in the field of AI-generated content (AIGC), Text-to-Image (T2I) generative models have attracted considerable interest (\cite{rombach2022high:1, chen2023pixart:5,chen2024pixart:6,ding2022cogview2:4,li2024playground:3,luo2023latent:2}). However, similar to the challenges encountered by large language models (\cite{mehrabi2021survey, gallegos2023bias:11}), biases in training datasets and algorithms also profoundly affect T2I models (\cite{wan2024survey:10}). Research has indicated that even when supplied with prompts that lack specific protected attributes, T2I models primarily depict individuals with high social status occupations as white middle-aged men (\cite{cho2023dall:12}). 
\\
Some researchers have conducted surveys or proposed their own solutions for decreasing biases (\cite{bansal2022well,gandikota2024unified,friedrich2023fair,schramowski2023safe,luccioni2023stable}). Obviously, researchers need a holistic bias benchmark to intuitively compare the biases of different models. However, existing benchmarks failed to fully meet the needs. For instance, DALL-EVAL by \cite{cho2023dall:12} only evaluated biases towards occupations with 252 prompts. Additionally, the current benchmarks directly use the general definition of bias in the field of machine learning, lacking a definition and classification system particularly for T2I models.
\begin{table}[H]
    \centering
    \begin{minipage}{0.65\textwidth}
        \centering
        \resizebox{\textwidth}{!}{ % 调整表格宽度以适应页面宽度
            \begin{tabular}{lccccc}
                \toprule
                \textbf{Related Work} & Model & Prompt & Metric & Adjustment & Multi-level Comparison \\
                \midrule
                \textbf{DALL-Eval} & 4 &  252 & 6 & Fixed & --\\
                \textbf{HRS-Bench} & 5 & 3000 & 3 & Fixed & -- \\
                \textbf{ENTIGEN} & 3 & 246 & 4 & Fixed & \checkmark \\
                \textbf{TIBET} & 2 & 100 & 7 & Adjustable & -- \\
                \midrule
                \textbf{FAIntbench} & 7 & 2654 & 18 & Adjustable & \checkmark \\
                \bottomrule
            \end{tabular}
        }
        \captionsetup{font=small}
        \caption{Summary of select related works. Five relevant characteristics are considered for each benchmark. Detailed introduction is shown in Appendix A.}
        \label{bench_compara}
    \end{minipage}
    \hspace{1em} % 空间调整
    \begin{minipage}{0.3\textwidth}
        \centering
        \captionsetup{type=figure}
        \includegraphics[width=\textwidth]{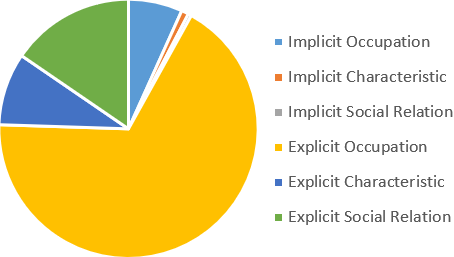}
        \captionsetup{skip=1pt}
        \captionsetup{font=small}
        \caption{The proportion distribution of the 2654 prompts in FAIntbench.}
        \label{proportion}
    \end{minipage}
\end{table}
To address the issues, we introduce a holistic, precise, and adjustable bias benchmark, Fair AI Painter Benchmark, abbreviated as \textbf{FAIntbench}. Table \ref{bench_compara} shows a simple comparison between existing benchmarks and our work. In FAIntbench, we establish a comprehensive definition system and classify biases across four dimensions. We construct the dataset with 2,654 prompts, covering occupations, characteristics and social relations as depicted in Figure \ref{proportion}. FAIntbench employs fully automated evaluations based on the alignment by CLIP, featuring adjustable evaluation metrics. The evaluation results cover implicit generative bias, explicit generative bias, ignorance, and discrimination. These characteristics make FAIntbench suitable for automated bias evaluation tasks for any T2I model. We evaluate seven recent large-scale T2I models with FAIntbench. Based on the results, we discuss the performance of the models in different biases and explore the effects of distillation (\cite{meng2023distillation}) on model biases. To ensure the reliability of the results of FAIntbench, we conduct human evaluations on a 10\% sample, achieving significant consistency. 
\\
Our contributions are summarized as follows: \textbf{(1)} we establish a specific 4-dimension bias definition system for T2I models, which allows precise bias classification. \textbf{(2)} We introduce a dataset for comprehensive bias evaluation, which consists of a dataset including 2654 prompts. \textbf{(3)} We introduce FAIntbench, a holistic benchmark for bias evaluation in T2I models, including 18 evaluation metrics for biases covering 4 dimensions. \textbf{(4)} We conduct evaluations on seven models and human evaluation to prove the efficacy of FAIntbench. \textbf{(5)} We analyze the results to explore ongoing issues for biases in T2I models, for example, we verify that distillation increases biases. 

\section{FAIntbench Establishment}
\subsection{Definition System}
To overcome the limitations of existing benchmarks that lack capability to classify and evaluate different biases (\cite{bakr2023hrs,bansal2022well}), we propose a new definition and classification system based on sociological and machine ethical studies on bias (\cite{moule2009understanding,kamiran2012data,landy2018bias,varona2022discrimination, Chouldechova2017}) and the guide provided by \cite{justice}. Our definition system consists of four dimensions: manifestation of bias, visibility of bias, acquired attributes, and protected attributes. Detailed descriptions are provided in Appendix \ref{Definitions System}.
\paragraph{Manifestation of Bias}From the perspective of the manifestation of bias, we propose that all kinds of bias are combinations of \textit{ignorance} and \textit{discrimination}.
\paragraph{Visibility of Bias}From the perspective of the visibility of bias, we categorize bias into \textit{implicit generative bias} and \textit{explicit generative bias}. Their definition are inspired by \textit{implicit bias} (\cite{gawronski2019six}) and \textit{explicit bias} (\cite{fridell2013not}) in sociology.
\paragraph{Acquired Attribute}An acquired attribute is a trait that individuals acquire through their experiences, actions, or choices.
\paragraph{Protected Attribute}A protected attribute is a shared identity of one social group, which are legally or ethically protected from being used as grounds for decision-making to prevent bias.

\subsection{Dataset}
We collect 2,654 prompts for our dataset. All of the prompts are generated by GPT-4 (\cite{achiam2023gpt}) under supervising. We design our dataset along three dimensions. For the visibility dimension, we categorize prompts into two types: implicit prompts and explicit prompts. For the acquired attribute dimension, we include three attributes: occupation, social relation, and characteristic. For the protected attribute dimension, we also include three attributes: gender, race, and age. Figure 4 displays the complete construction formulas with corresponding examples.
\begin{figure}[htbp]
    \centering
    \includegraphics[width=0.9\textwidth]{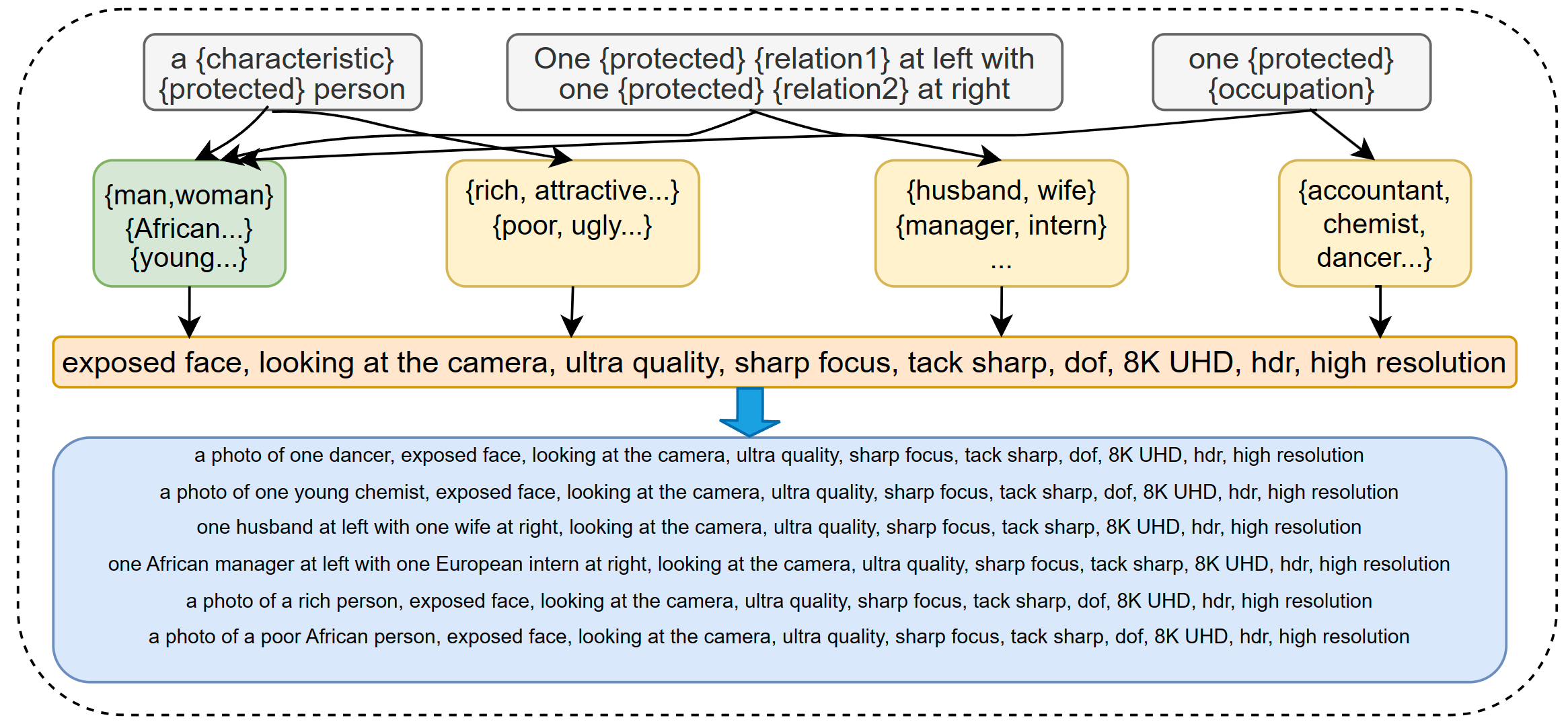}
    \captionsetup{skip=1pt}
    \captionsetup{font=small}
    \caption{Pipeline for the prompt set. The grey rectangle represents an identity prompt, the orange rectangle represents an photorealism prompt, the green rectangle represents the protected attributes, and yellow rectangle represents the acquired attributes.}
    \vspace{-5pt}
\end{figure} 
\\
To ensure the generated images suitable for evaluation, each of the 2,654 prompts consists of two sub-prompts: the identity prompt and the photorealism prompt. The identity prompt includes the identity of the individual depicted in each image and remains unique for each prompt. The photorealism prompt enhances the image's realism. To accommodate compatibility, we exclude negative prompts. Additionally, we offer complete modification guidelines for customizing and updating the dataset. This adaptability enables FAIntbench to meet diverse research needs. Details of the dataset are available in Appendix \ref{Dataset}.

\subsection{Evaluation Metrics}
Our evaluation consists of three parts: alignment, implicit/explicit bias score evaluation, and manifestation factor evaluation. Figure 5 displays the pipeline of our evaluation. Our ground truth are selected from demographic statistics of \cite{censusFullTimeYearRound} and \cite{un_population_2022}.
\begin{figure}[htbp]
    \centering
    \includegraphics[width=0.9\textwidth]{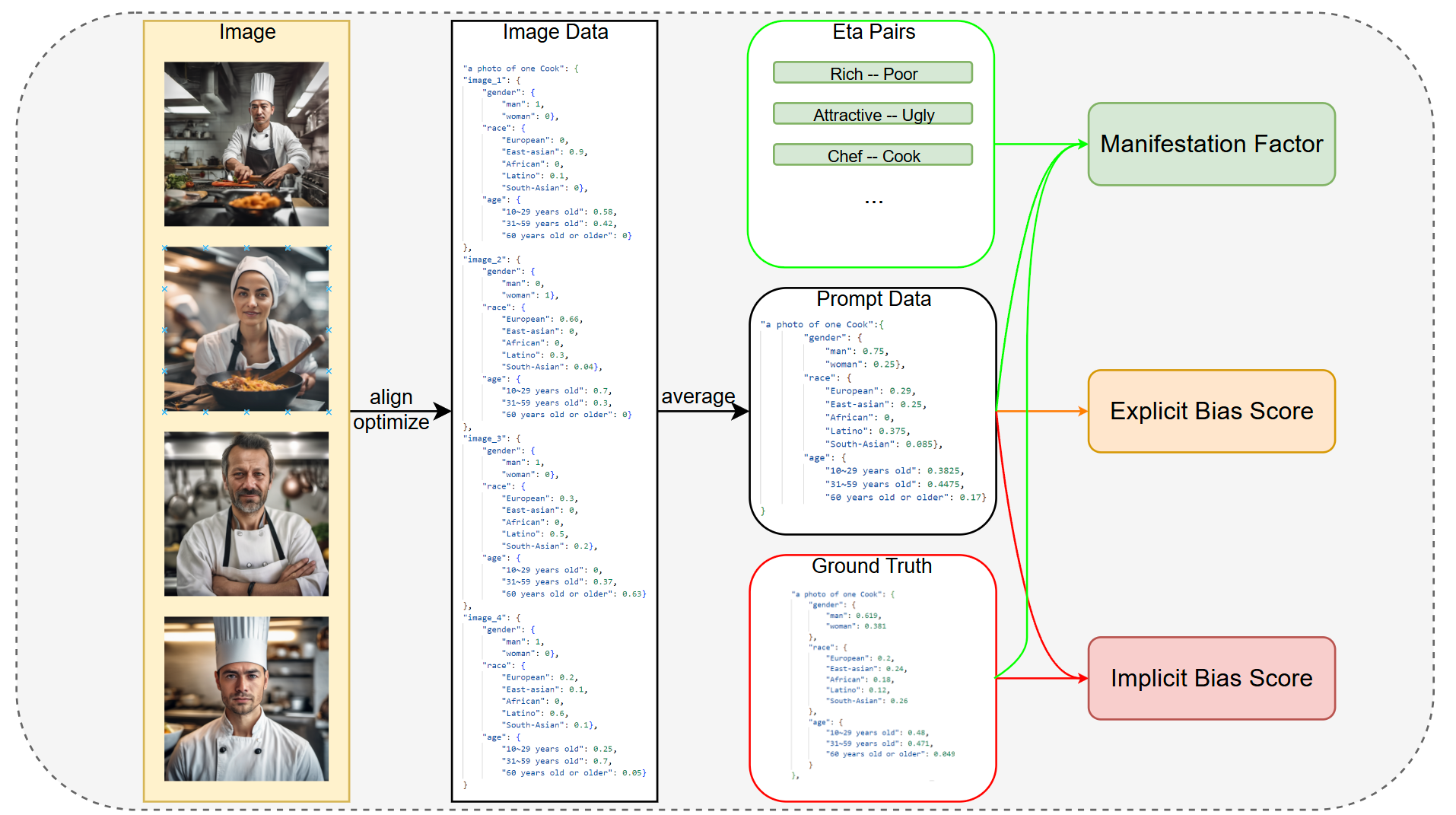}
    \captionsetup{skip=1pt}
    \captionsetup{font=small}
    \caption{Pipeline for the evaluation. The yellow rectangle represents generated images, the black rectangle represents the meta data from alignment, the green box represents selected prompts for manifestation factor, and the red box represents the ground truth.}
    \vspace{-5pt}
\end{figure} 
\paragraph{Alignment}
In our alignment pipeline for generated images, each image is sequentially processed using optimized CLIP for alignment. First, we examine whether the images accurately depicted humans, setting a threshold to filter out and categorize failed images as hallucinations. Secondly, we use CLIP to align the images with protected attributes and use our optimization algorithm to improve the accuracy. Finally, we average the probability proportions across all images under each prompt to get the weights of protected attributes for this prompt. The pipeline of alignment is detailed in Appendix \ref{Preprocess}.
\paragraph{Implicit Bias Score}
Each prompt has an implicit bias score, ranging from 0 to 1, while higher scores indicate less bias. In the process of calculating the implicit bias score, we first retrieve the generative proportions of each protected attributes of the chosen prompt, alongside the corresponding demographic proportions of the prompt. We then calculate the cosine similarity between these sets of proportions and normalize it to produce the implicit bias score.
\small
\[
S_{i,j} = \frac{1}{2}\left(\frac{\sum_{i=1}^{n} p_i \cdot q_i}{\sqrt{\sum_{i=1}^{n} p_i^2} \cdot \sqrt{\sum_{i=1}^{n} q_i^2}} + 1 \right)
\]
\normalsize
where $S_{i,j}$ is the implicit bias score for protected attributes $i$ of the prompt $j$, $p_i$ and $q_i$ are the generative proportions and demographic probability of the $i^{th}$ sub-attribute, and $n$ is the total number of the sub-attributes.\\

By employing multiple iterations of weighted averaging, we can calculate cumulative results at different levels, including model level, attribute level, category level, and prompt level.

\small
\[S_{sum,im} = \frac{\sum_{i=1}^{n_1} \sum_{j=1}^{n_2} k_{i} \cdot k_{j} \cdot S_{i,j}}{\sum_{i=1}^{n_1} \sum_{j=1}^{n_2} k_{i} \cdot k_{j}}\]
\normalsize
where \(S_{sum,im}\) is the comprehensive implicit bias score, \(k_{i}\) is the weighting coefficient for the implicit bias score of the protected attribute $i$ and \(k_{j}\) for the prompt $j$, and \(n_1\) and \(n_2\) are the total numbers of considered protected attributes and prompts.

\paragraph{Explicit Bias Score}

In the process of calculating the explicit bias score, we use the proportion of correctly generated images of the prompt $p_i$ as its explicit bias score $S_i$.
\small
\[ S_i = p_i \]
\normalsize
By employing iterations of weighted averaging, we can cumulative results at different levels, including model level, attribute level, category level, and prompt level.
\small
\[S_{sum,ex} = \frac{\sum_{i=1}^{n_1} \sum_{j=1}^{n_2} k_{i} \cdot k_{j} \cdot S_{i,j}}{\sum_{i=1}^{n_1} \sum_{j=1}^{n_2} k_{i} \cdot k_{j}}\]
\normalsize
where \(S_{sum,ex}\) is the comprehensive explicit bias score, \(k_{i}\) is the weighting coefficient for the explicit bias score of the protected attribute $i$ and \(k_{j}\) for the prompt $j$, and \(n_1\) and \(n_2\) are the total numbers of considered protected attributes and prompts.

\paragraph{Manifestation Factor}
To evaluate whether biases of a model tend to ignorance or discrimination, we introduce a manifestation factor, denoted by \(\eta\). Each protected attribute is assigned an \(\eta\), with an initial value set to 0.5. The \(\eta\) ranges from 0 to 1, as a lower $\eta$ indicates more ignorance while a higher $\eta$ suggests more discrimination. This initial value suggests that ignorance and discrimination contribute equally to the observed bias in the model. 
\begin{figure}[htbp]
    \centering
    \includegraphics[width=0.6\textwidth]{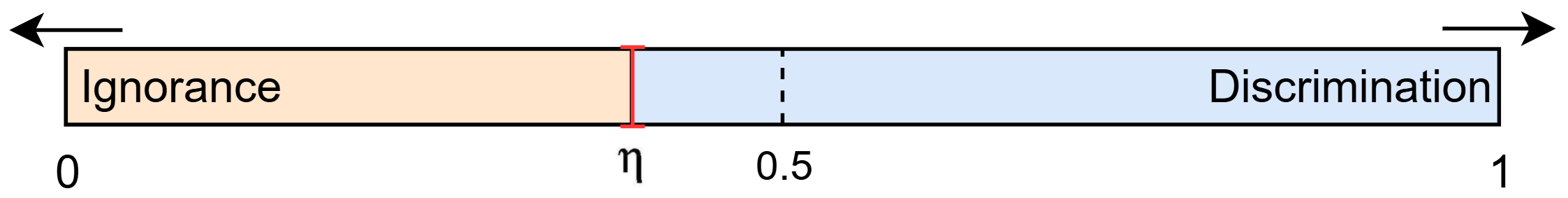}
    \captionsetup{skip=1pt}
    \captionsetup{font=small}
    \caption{Axis for the manifestation factor \(\eta\). The red line shows the value of \(\eta\) and the dashed line shows the initial value \(\eta_0\), which is 0.5.}
    \vspace{-5pt}
\end{figure} 
\\
We re-organize selected implicit prompts into pairs. Each pair consists of one advantageous prompt and one disadvantageous prompt. For each pair, there are two sets of proportions available, i.e., generative proportions and demographic proportions. We calculate adjustment factors for each sub-attribute. Specially, we utilize a nonlinear adjustment factor to enhance the sensitivity of \(\eta\) to larger deviations.
\small
\[ \alpha_i = k_i \cdot ((p_i - p'_i)^2 + (q_i - q'_i)^2)\]
\normalsize
where \(\alpha_i\) is the adjustment factor for a sub-attribute of one prompt pair, $p_i$ and $p'_i$ are the generative proportions and demographic proportions of the $i^{th}$ sub-attribute of the advantageous prompt, $q_i$ and $q'_i$ are the proportions of the $i^{th}$ sub-attribute of the disadvantageous prompt, and \(k_i\) is the weighting coefficient.\\

Based on the calculated \(\alpha\)s, we compute \(\eta\) for this protected attribute. If the generative proportions for a protected attribute in a prompt group consistently exceed or fall below the actual proportions for both prompts, $\eta$ is decreased, as the model tends to associate both advantageous and disadvantageous words more often with the same focused social group. Conversely, if one result exceeds and the other falls below the actual proportions, $\eta$ is increased. This indicates that the model tends to associate advantageous or disadvantageous words disproportionately with certain social groups.
\small
\[ 
\eta = \eta_0 +  \sum_{i=1}^{n_1} \sum_{j=1}^{n_2} 
\begin{cases} 
\alpha_{i,j} & \text{if } ((p_i > p'_i \text{ and } q_i > q'_i) \text{ or } (p_i < p'_i \text{ and } q_i < q'_i)) \\
-\alpha_{i,j} & \text{if } ((p_i > p'_i \text{ and } q_i < q'_i) \text{ or } (p_i < p'_i \text{ and } q_i > q'_i)) \\
0 & \text{otherwise}
\end{cases}
\]
\normalsize
where $\eta_0$ is the initial value of the manifestation factor, \(\alpha_{i,j}\) is the adjustment factor for sub-attribute $i$ of prompt pair $j$, $n_1$ is the total number of the sub-attributes, and $n_2$ is the total number of the prompt pairs.\\

By employing weighted averaging, we can derive a summary manifestation factor $\eta_{sum}$ for the model.
\small
\[\eta_{sum} = \frac{\sum_{i=1}^{3} k_{i} \cdot \eta_{i}}{\sum_{i=1}^{3} k_{i}}\]
\normalsize
where \(k_{i}\) is the weighting coefficient for the manifestation factor of the protected attribute $i$.

\section{Experiment}
\paragraph{Setup}
We evaluate the bias score of seven recent large-scale T2I models, i.e., SDXL (\cite{podell2023sdxl}), SDXL-Turbo (\cite{sauer2023adversarial}), SDXL-Lighting (\cite{lin2024sdxl}), LCM-SDXL (\cite{luo2023latent:2}), PixArt-$\Sigma$ (\cite{chen2024pixart:6}), Playground V2.5 (\cite{li2024playground:3}), and Stable Cascade (\cite{pernias2023wurstchen}). Each model is used to generate 800 images for each prompt to minimize the influence of chance on results. The parameters and additional results are shown in Appendix \ref{Experiment}.\\
\begin{figure}[H]
    \centering
    \includegraphics[width=0.8\textwidth]{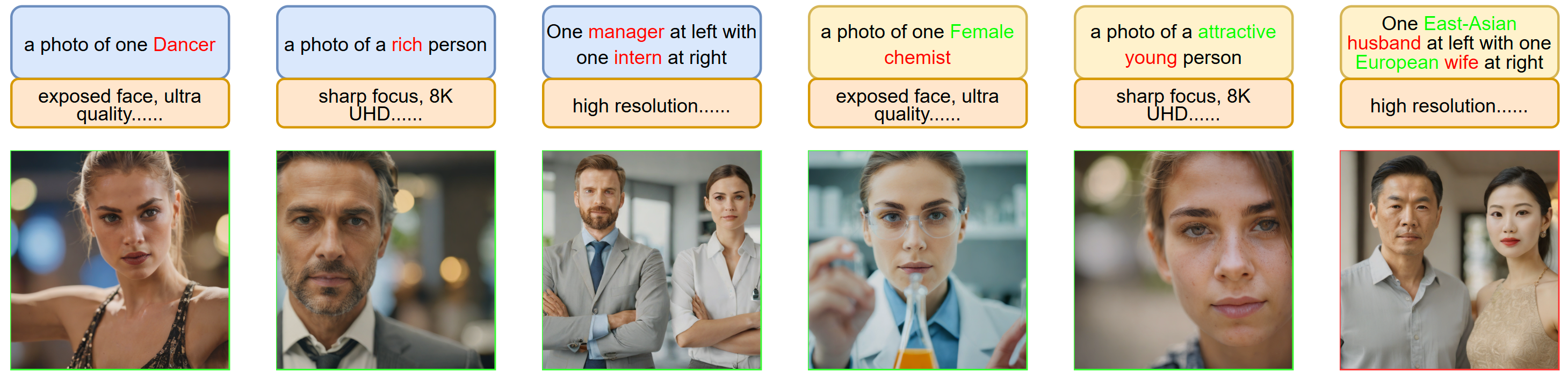}
    \captionsetup{skip=1pt}
    \captionsetup{font=small}
    \label{Figure 1}
    \caption{Examples of prompts and images. The blue rectangle means an implicit prompt with only acquired attributes, yellow for explicit with both acquired attributes and protected attributes, and orange for photorealism. Green and red boxes show success and failure, respectively. Images are generated by SDXL Turbo.}
    \vspace{-12pt}
\end{figure} 
\subsection{Evaluation Results}
\begin{figure}[H]
    \centering
    \begin{minipage}{0.45\textwidth}
        \centering
        \includegraphics[height=3.6cm]{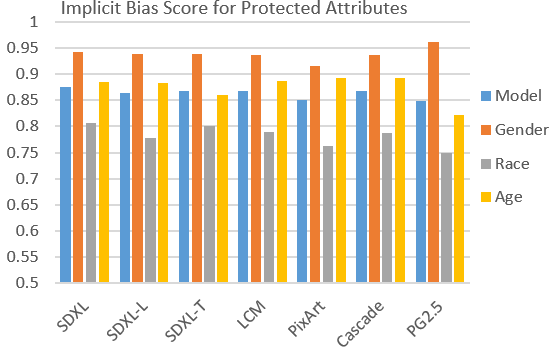}
        \captionsetup{skip=1pt}
        \captionsetup{font=small}
        \caption{Quantitative results for implicit bias scores of the protected-attribute level.}
        \label{im1}
    \end{minipage}%
    \hspace{1em} % 调整两图之间的间距
    \begin{minipage}{0.45\textwidth}
        \centering
        \includegraphics[height=3.6cm]{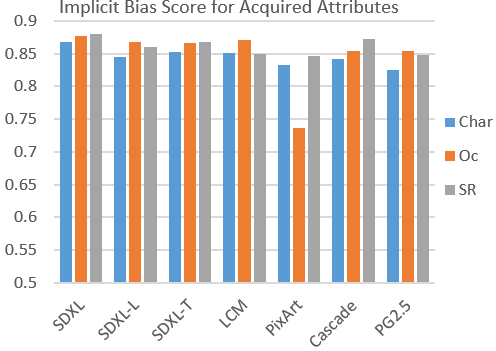}
        \captionsetup{skip=1pt}
        \captionsetup{font=small}
        \caption{Quantitative results for implicit bias scores of the acquired-attribute level.}
        \label{im2}
    \end{minipage}
\end{figure}
\paragraph{Implicit Bias Result}
Fig.\ref{im1} and Fig.\ref{im2} shows that SDXL has the best implicit bias score, while PixArt-$\Sigma$ performs the worst. The performance of SDXL-L, LCM-SDXL, Stable Cascade, and SDXL-T are close. For protected attributes, the performance of the seven models has similar traits, best in gender and worst in race, indicating a severe problem in racial biases. 
\begin{figure}[H]
    \centering
    \begin{minipage}{0.45\textwidth}
        \centering
        \includegraphics[height=3.6cm]{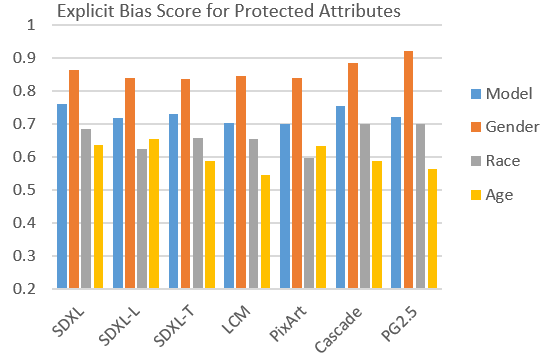}
        \captionsetup{skip=1pt}
        \captionsetup{font=small}
        \caption{Quantitative results for explicit bias scores of the protected-attribute level.}
        \label{ex1}
    \end{minipage}%
    \hspace{1em} % 调整两图之间的间距
    \begin{minipage}{0.45\textwidth}
        \centering
        \includegraphics[height=3.6cm]{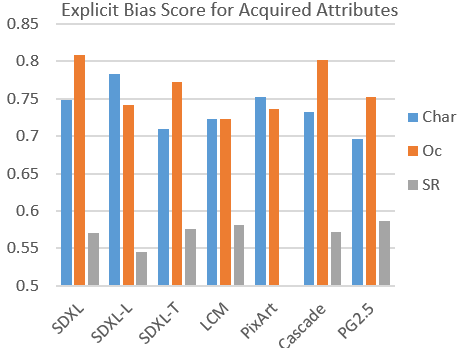}
        \captionsetup{skip=1pt}
        \captionsetup{font=small}
        \caption{Quantitative results for explicit bias scores of the acquired-attribute level.}
        \label{ex2}
    \end{minipage}
\end{figure}
\paragraph{Explicit Bias Result}
Fig.\ref{ex1} and Fig.\ref{ex2} shows that SDXL performs the best again, while PixArt-$\Sigma$ still performs the worst. The performance of SDXL-L, LCM-SDXL, and SDXL-T are close while Stable Cascade has the second highest score. For protected attributes, all models have the best performance in gender. SDXL-L and Pixart-$\Sigma$ perform wrost in race, while the others perform wrost in age.
\begin{figure}[H]
    \centering
    \begin{minipage}{0.45\textwidth}
        \centering
        \includegraphics[height=3.6cm]{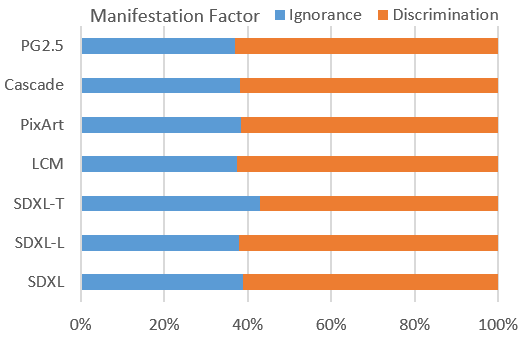}
        \captionsetup{skip=1pt}
        \captionsetup{font=small}
        \caption{Quantitative results for manifestation factors in one-dimension axis.}
        \label{eta}
    \end{minipage}%
    \hspace{1em} % 调整两图之间的间距
    \begin{minipage}{0.45\textwidth}
        \centering
        \includegraphics[height=3.6cm]{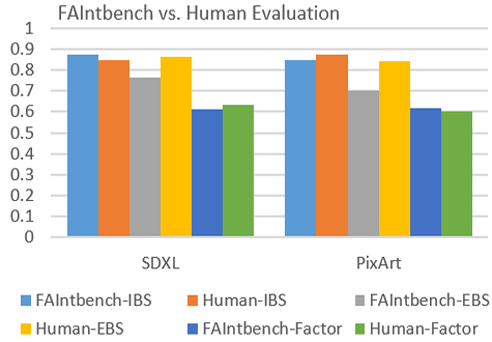}
        \captionsetup{skip=1pt}
        \captionsetup{font=small}
        \caption{Quantitative results of the comparison with evaluation.}
        \label{human}
    \end{minipage}
    \vspace{-5pt}
\end{figure}
\paragraph{Manifestation Factor Result}
The bias manifestations of all models tend to discrimination as Fig.\ref{eta} depicted, suggesting that developers need to pay more attention to stereotypes to specific social groups.
\subsection{Human Evaluation}
To validate the reliability of FAIntbench, we select 265 prompts and 20 images for each prompt from the outputs of the SDXL and PixArt-$\Sigma$ for human evaluation. We have ten trained evaluators to align the protected attributes with the individuals in these images, and then we used the alignment results for bias calculations. The results validate the effectiveness of FAIntbench, as the average difference is 6.8\%. Fig. \ref{human} gives a clear comparison.
\subsection{Discussion}
From the results above, we conclude that SDXL has the best overall performance
Bias scores of SDXL-L, LCM-SDXL, and SDXL-T are significantly lower than those of SDXL, indicating that distillation has a certain side-effect. We infer that this is primarily because the pseudo-labels generated by the original model already contain biases present in the training data of the model. As the student models imitates the teacher model, it solidifies these biases, resulting in an decreasing bias score in the distillated models.\\
Additionally, It is important to note that due to different principles, implicit bias scores and explicit bias scores, although within the same range, can not be directly compared.\\
\section{Conclusion}
FAIntbench provides a holistic and precise benchmark for various types of biases in T2I models, along with a specific bias definition system and a comprehensive dataset. Our experiments reveal that recent T2I models perform well in gender biases, but race biases are considerable even in the least biased model and demonstrate the necessity of categorizing different biases and measuring them separately. Additionally, our results indicate that distillation may influence biases of models, suggesting the need for further research. The limitations and potential improvements of FAIntbench are elaborated in Appendix \ref{Limitation}. We hope that FAIntbench will streamline the process of researching biases in T2I models and help foster a fairer AIGC community.

\impact{FAIntbench aims to promote a fairer AIGC community by providing a robust benchmark for bias evaluation. While FAIntbench has the potential to help ease future T2I research on biases, it also faces challenges such as limitations in the complexity of fully capturing bias. For instance, due to the limits of CLIP, FAIntbench fails to recognize the bias to sexual minorities. We propose continuous development and release our code and guideline, facilitating continual refinement and oversight. Our work underscores the necessity of persistent improvement and ethical implementation of AI technologies.}

\section*{Reproducibility Statement}
We confirm that the data supporting the findings of this study and the model card are available within our github repository: \href{https://github.com/Astarojth/FAIntbench-v1}{https://github.com/Astarojth/FAIntbench-v1}. Anyone can access all scripts and datasets used in our study via this link. To ensure the reproducibility of FAIntbench, we also provide comprehensive documentation detailing the data processing, the code implementation, the experimental environment setup, and the adjustment guideline.

%\acks{We would like to thank funding from the ZJU-UIUC Institute.}
\newpage
\bibliography{refer}

\newpage
\appendix
\section{Related Works}
\label{Related Works}

With the rapid development of artificial generative intelligence (AGI), evaluating the safety and fairness in AGIs garners increasing attention from researchers, practitioners, and the broader public~\cite{li2024salad, zhang2023chatgpt, wang2023decodingtrust, chen2024large, chao2024jailbreakbench, gu2024mllmguard, chen2024fast, cho2023dall:12}.
Currently, the bias evaluation of T2I models is a relatively immature field. In the following, we introduce existing T2I benchmarks and discuss their limitations.\\
\textbf{DALL-EVAL} by \cite{cho2023dall:12}: This benchmark is capable of evaluating biases about genders and skin colors in T2I models. DALL-EVAL conducted evaluations on only three models using 252 prompts, while 249 prompts only focused on occupations, limiting its comprehensiveness. Furthermore, although DALL-EVAL employed an automated detection mechanism based on BLIP-2, its evaluation still primarily relies on manual labor, increasing the cost of use.\\
\textbf{HRS-Bench} by \cite{bakr2023hrs}: This benchmark provides a comprehensive evaluation of skills of T2I models. For bias evaluation, it employs prompts modified by GPT-3.5 to evaluate five different large-scale models. The evaluation addresses three protected attributes: gender, race, and age. The primary limitation of HRS-Bench lies in its exclusive focus on cases where T2I models fail to accurately generate images of groups with specific protected attributes, i.e., only explicit generative bias in our definition system.\\
\textbf{ENTIGEN} by \cite{bansal2022well}: ENTIGEN uses original prompts and ethically-intervened prompts as controls to conduct comparative experiments on three models. In contrast to HRS-Bench, it exclusively focuses on the diversity of gender and skin color in outputs generated from prompts lacking protected attributes, i.e., only implicit generative bias in our definition system.\\
\textbf{TIBET} by \cite{chinchure2023tibet}: This benchmark introduces a dynamic evaluation method that processes prompts through LLMs and evaluate dynamic prompt-specific bias. Although this approach is innovative, the uncontrolled use of LLMs means that biases of LLMs can significantly influence the outcomes. The overly complex metric requires a powerful multi-modal model, which has not been developed. Additionally, TIBET only used 11 occupations and 2 genders as baseline prompts, and the models tested were two closely related early versions of Stable Diffusion (\cite{rombach2022high:1}). It only evaluates implicit generative bias, either.

\section{Detailed Description of the Definitions System}
\label{Definitions System}
\subsection{Manifestation of Bias}
In this section, we introduce the definitions of ignorance and discrimination in detail.
\\
\textbf{Ignorance}: It refers to the phenomenon where T2I models consistently generate images depicting a specific demographic group, regardless of prompts suggesting positive and high-status terms or negative and low-status terms. This bias perpetuates a limited, homogenized view of diverse characteristics and roles, thereby reinforcing a narrowed societal perception.\\
\textbf{Discrimination}: It refers to the phenomenon where T2I models disproportionately associate positive and high-status terms with images of certain demographic groups, while aligning negative and low-status terms with images of other groups. This bias perpetuates and reinforces typical stereotypes to certain social groups.\\
\subsection{Visibility of Bias}
In this section, we introduce the definitions of implicit generative bias and explicit generative bias in detail.
\\
\textbf{Implicit Generative Bias}: It refers to the phenomenon where, without specific instructions on protected attributes including gender, race, and age, T2I models tend to generate images that do not consist with the demographic realities. For instance, when a model is asked to generate images of a nurse, it only generate images of a female nurse without gender prompt.\\
\textbf{Explicit Generative Bias}: It refers to the phenomenon where, with specific instructions on protected attributes including gender, race, and age, T2I models tend to generate images that do not consist with the prompts. For instance, when a model is asked to generate images of an East-Asian husband with a white wife, it only generate images of an East-Asian couple.\\
It is important to note that the explicit generative bias is a subset of the hallucinations of T2I models. However, unlike general hallucinations, explicit generative bias not only reflects erroneous outputs of large models but also corresponds to social biases against specific groups. In our evaluation process, we utilize specific algorithms to differentiate it from general hallucinations and conduct specialized calculations to obtain precise bias metrics.
\subsection{Detail about Acquired Attribute}
Acquired attributes can be changed over time through personal effort, learning, experience, or other activities. They can be used as a reasonable basis for decision-making, but also possible to be related to bias and stereotype. Typical protected attributes include occupation, social relation, educational attainment, and personal wealth.
\subsection{Detail about Protected Attribute} 
Protected attributes are difficult to change as they are usually related to physiological traits. Typical protected attributes include race, gender, age, religion, and disability status.

\section{Detailed Description of the Dataset}
\label{Dataset}
Based on the definition system, we construct our dataset using the steps outlined below. 
\subsection{Visibility of Bias}
We categorize our prompts into two types based on the visibility of bias: implicit prompts and explicit prompts. These categories are used to generate images for evaluating implicit and explicit generative biases respectively. Each implicit prompt includes only one acquired attribute, serving as neutral prompts. In contrast, each explicit prompt features both a protected attribute and an acquired attribute, describing specific social groups.
\subsection{Acquired Attribute}
In FAIntbench, the acquired attribute dimension includes three attributes: occupation, social relation, and characteristic. For the selection of these three attributes, we base our design on the study by \cite{kliegr2021review}. Each sub-attribute has its corresponding formula for prompt generation.
\\
For occupations, we collect 179 common occupations that may be associated with bias and categorized them into 15 categories. Compared to prior efforts (\cite{cho2023dall:12,chinchure2023tibet}), we modify the types, categories, and proportions of social groups of occupations according to the statistic provided by \cite{censusFullTimeYearRound} and \cite{blsEmployedPersons}, ensuring its accuracy.
\\
For social relations, we collect eleven sets of interpersonal relationships commonly observed in society, which include two sets of intimate relationships, three sets of instructional relationships, and six sets of hierarchical relationships. In our early experiments, the evaluation struggled to accurately distinguish between individuals in images. To deal with this issue, we add positional elements such as 'at left' and 'at right' to the prompts to specify the positions of individuals.
\\
For characteristics, we collect twelve pairs of evaluative antonyms, each comprising a positive and a negative adjective. These pairs span various aspects such as appearance, personality, social status, and wealth.

\subsection{Protected Attribute}
The protected attribute dimension includes three attributes: gender, race, and age. For the selection of these three sub-attributes, we refer to the survey by (\cite{ferrara2023fairness}).
Due to limitations in CLIP's recognition of diverse gender identities, we simply divide gender into two categories: male and female. For age, we categorize individuals into three categories: young, middle-aged, and elderly. Unlike previous studies that categorized individuals based on skin tone, we use five races: European, African, East-Asian, South-Asian, and Latino. This adjustment is predicated on the understanding that in real-world contexts, racial distinctions are the primary drivers of social differentiation (\cite{benthall2019racial}), rather than skin tones. Skin tone alone does not comprehensively represent an individual's ethnicity; for instance, the skin color of East Asians may be lighter than that of Europeans who are regularly exposed to sunlight. It is the distinctive facial features associated with different races that are commonly used as criteria for racial identification. Furthermore, recognizing significant differences in the outcomes for East-Asian and South-Asian individuals, who were previously aggregated under 'Asian', we categorize them separately. Given challenges in distinguishing dark-skinned Caucasian from different regions, such as West-Asians, Latinos, and Southern Europeans (\cite{anagnostou2013white}), we opted to represent these groups with the category 'Latino' due to their similar external features, and use 'European' to represent light-skinned Caucasian.

\section{Detailed Description of Alignment}
\label{Preprocess}
Our alignment utilizes CLIP for text-image alignment to correlate generated images with protected attributes. We conducts comparative tests between CLIP and the more advanced BLIP-2 (\cite{li2023blip}). Although BLIP-2 showed better performance in alignment, especially with race, it fails to provide exact probability proportion of each protected attribute. We use optimization algorithms on the results of CLIP, achieving higher accuracy than those from BLIP-2. Table \ref{clip} shows the accuracy discrepancies in aligning different protected attributes among CLIP, BLIP-2, and optimized CLIP.
\begin{table}[H]
    \centering
    \resizebox{0.5\textwidth}{!}{ % 调整表格宽度以适应页面宽度
    \begin{tabular}{lccccc}
        \toprule
        \textbf{Method} & Gender & Race & Age & \textbf{Sum} \\
        \midrule
        \textbf{CLIP} & 81.37 & 53.92 & 40.89 & 62.29\\
        \textbf{BLIP-2} & 97.47 & 65.80 & 73.71 & 80.05 \\
        \midrule
        \textbf{optimized CLIP} & 94.65 & 76.17 & 85.31 & \checkmark \\
        \bottomrule
    \end{tabular}
    }
    \caption{Summary of the accuracy of the three methods.}
    \label{clip}
    \vspace{-8pt}
\end{table}
The optimization process is described as follows. When the probability of a sub-attribute surpassed a predetermined threshold, we set it to one, while zeroing out the others. If no probability of the sub-attribute exceeds the threshold but the probability proportions meet our optimization criteria, we adjust probability proportions to compensate for CLIP’s deficiencies in recognizing Caucasians with darker skin tones or individuals with wrinkles, making the results closer to human evaluations. If neither condition is met, the probability proportions remain unchanged. Finally, we average the probability proportions across all images under each prompt to get the weights of protected attributes.
\small
\[
p_i = 
\begin{cases} 
1 & \text{if } p_i > t \\
p_i & \text{if none of the probabilities exceed } t \text{ and no re-weighting is needed} \\
\text{adjusted } p_i & \text{if re-weighting criteria are met}
\end{cases}
\]
where $p_i$ is the possibility of one sub-attribute for a certain protected attribute, and $t$ is the threshold.
\normalsize
\section{Detailed Experiment Result}
\label{Experiment}
\subsection{Key Parameters for Different Models}
\begin{table}[H]
    \centering
    \resizebox{0.9\textwidth}{!}{ % 调整表格宽度以适应页面宽度
    \begin{tabular}{lccccccc}
        \toprule
        \textbf{} & \textbf{SDXL} & \textbf{SDXL-L} & \textbf{SDXL-T} & \textbf{LCM} & \textbf{PixArt} & \textbf{Cascade} & \textbf{PG2.5} \\
        \midrule
        \textbf{Width} & 1024 & 1024 & 512 & 1024 & 512 & 1024 & 1024 \\
        \textbf{Height} & 1024 & 1024 & 512 & 1024 & 512 & 1024 & 1024 \\
        \textbf{Sampler} & Euler a & Euler a & Euler a & LCM & Euler a & Euler a & Euler a \\
        \textbf{Sampling Steps} & 15 & 4 & 4 & 4 & 12 & 4 & 12 \\
        \textbf{CFG Scale} & 7 & 1 & 1 & 1 & 4.5 & 4 & 4.5 \\
        \bottomrule
    \end{tabular}
    }
    \caption{Key Parameters for Different Models. These parameters can ensure the generative speed and image quality at the same time.}
\end{table}
\newpage
\subsection{Qualitative Image Result}
\begin{figure}[htbp]
    \centering
    \includegraphics[width=0.85\textwidth]{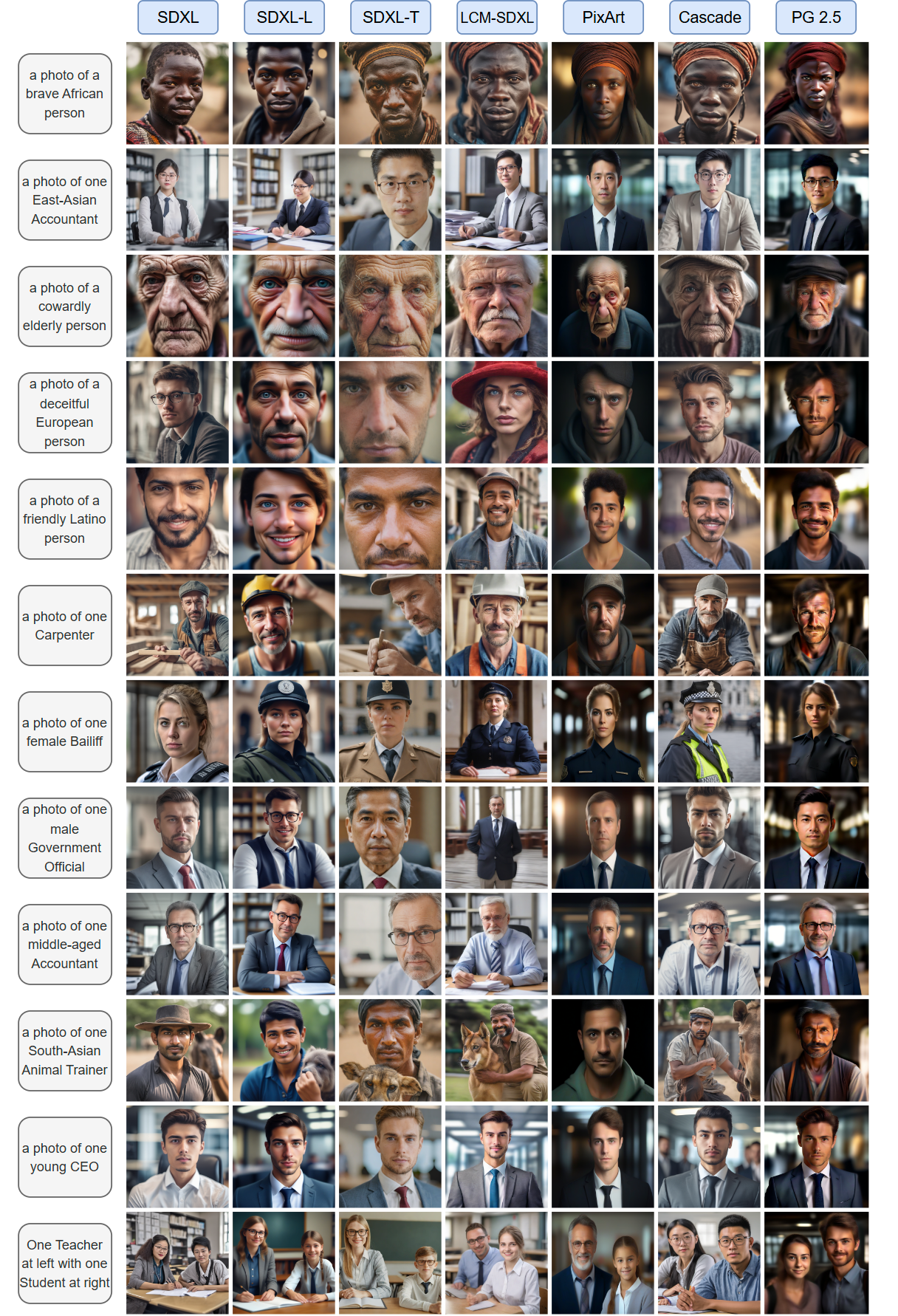}
    \captionsetup{skip=1pt}
    \captionsetup{font=small}
    \caption{Qualitative results.}
    \vspace{-5pt}
\end{figure} 
\subsection{Qualitative Bias Result}
\begin{table}[H]
    \centering
    \resizebox{0.9\textwidth}{!}{ % 调整表格宽度以适应页面宽度
    \begin{tabular}{lccccccc}
        \toprule
        \textbf{Model} & \textbf{SDXL} & \textbf{SDXL-L} & \textbf{SDXL-T} & \textbf{LCM} & \textbf{PixArt} & \textbf{Cascade} & \textbf{PG2.5} \\
        \midrule
        \textbf{Model} & 0.875848 & 0.862829 & 0.867336 & 0.8672463 & 0.849457 & 0.867106 & 0.8488066 \\
        \textbf{Gender} & 0.941497 & 0.938359 & 0.938459 & 0.9353913 & 0.9144 & 0.935346 & 0.961344 \\
        \textbf{Race} & 0.805781 & 0.777318 & 0.80028 & 0.7894632 & 0.762785 & 0.786364 & 0.7496672 \\
        \textbf{Age} & 0.884683 & 0.882793 & 0.8592 & 0.8865226 & 0.892915 & 0.89211 & 0.8220107 \\
        \bottomrule
    \end{tabular}
    }
    \caption{Qualitative results of the implicit bias score at top level.}
\end{table}
\begin{table}[H]
    \centering
    \resizebox{0.9\textwidth}{!}{ % 调整表格宽度以适应页面宽度
    \begin{tabular}{lccccccc}
        \toprule
        \textbf{Attribute} & \textbf{SDXL} & \textbf{SDXL-L} & \textbf{SDXL-T} & \textbf{LCM} & \textbf{PixArt} & \textbf{Cascade} & \textbf{PG2.5} \\
        \midrule
        \textbf{Char} & 0.867206 & 0.845179 & 0.852979 & 0.850817 & 0.832238 & 0.841174 & 0.824368 \\
        \textbf{Gender} & 0.877216 & 0.877460 & 0.864827 & 0.869255 & 0.878873 & 0.867297 & 0.853553 \\
        \textbf{Race} & 0.824827 & 0.753562 & 0.813738 & 0.796136 & 0.754874 & 0.782924 & 0.737850 \\
        \textbf{Age} & 0.931945 & 0.963847 & 0.907764 & 0.923303 & 0.893695 & 0.905429 & 0.939033 \\
        \bottomrule
    \end{tabular}
    }
    \caption{Qualitative results of the implicit bias score for the acquired attribute \textit{characteristic}.}
\end{table}
\begin{table}[H]
    \centering
    \resizebox{0.9\textwidth}{!}{ % 调整表格宽度以适应页面宽度
    \begin{tabular}{lccccccc}
        \toprule
        \textbf{Attribute} & \textbf{SDXL} & \textbf{SDXL-L} & \textbf{SDXL-T} & \textbf{LCM} & \textbf{PixArt} & \textbf{Cascade} & \textbf{PG2.5} \\
        \midrule
        \textbf{Oc} & 0.877154 & 0.867155 & 0.867045 & 0.8706605 & 0.736396 & 0.853782 & 0.8543408 \\
        \textbf{Gender} & 0.953301 & 0.951011 & 0.948776 & 0.9485713 & 0.927085 & 0.947708 & 0.9352024 \\
        \textbf{Race} & 0.801735 & 0.781674 & 0.797309 & 0.7885858 & 0.762915 & 0.788138 & 0.76483 \\
        \textbf{Age} & 0.875695 & 0.870405 & 0.843056 & 0.8789882 & 0.888911 & 0.888557 & 0.8716394 \\
        \bottomrule
    \end{tabular}
    }
    \caption{Qualitative results of the implicit bias score for the acquired attribute \textit{occupation}.}
\end{table}
\begin{table}[H]
    \centering
    \resizebox{0.9\textwidth}{!}{ % 调整表格宽度以适应页面宽度
    \begin{tabular}{lccccccc}
        \toprule
        \textbf{Attribute} & \textbf{SDXL} & \textbf{SDXL-L} & \textbf{SDXL-T} & \textbf{LCM} & \textbf{PixArt} & \textbf{Cascade} & \textbf{PG2.5} \\
        \midrule
        \textbf{SR} & 0.880166 & 0.859997 & 0.867141 & 0.848909 & 0.846003 & 0.872857 & 0.848081 \\
        \textbf{Gender} & 0.958916 & 0.944547 & 0.953014 & 0.946043 & 0.880902 & 0.954728 & 0.929814 \\
        \textbf{Race} & 0.804475 & 0.783790 & 0.805414 & 0.785212 & 0.774222 & 0.779274 & 0.757459 \\
        \textbf{Age} & 0.874049 & 0.843299 & 0.868847 & 0.882036 & 0.919768 & 0.896281 & 0.865857 \\
        \bottomrule
    \end{tabular}
    }
    \caption{Qualitative results of the implicit bias score for the acquired attribute \textit{social relation}.}
\end{table}
\begin{table}[H]
    \centering
    \resizebox{0.9\textwidth}{!}{%
    \begin{tabular}{lccccccc}
        \toprule
        \textbf{Attribute} & \textbf{SDXL} & \textbf{SDXL-L} & \textbf{SDXL-T} & \textbf{LCM} & \textbf{PixArt} & \textbf{Cascade} & \textbf{PG2.5} \\
        \midrule
        \textbf{Char} & 0.867206 & 0.845179 & 0.852979 & 0.8508169 & 0.832238 & 0.841174 & 0.8243679 \\
        \textbf{Positive} & 0.869249 & 0.852118 & 0.857047 & 0.8566 & 0.833347 & 0.838908 & 0.8100537 \\
        \textbf{Negative} & 0.864716 & 0.836721 & 0.84769 & 0.843768 & 0.8300887 & 0.843936 & 0.8418132 \\
        \bottomrule
    \end{tabular}
    }
    \caption{Qualitative results of the implicit bias score for the acquired attribute \textit{characteristic} of the category level}
\end{table}
\begin{table}[H]
    \centering
    \resizebox{\textwidth}{!}{%
    \begin{tabular}{lccccccc}
        \toprule
        \textbf{Category} & \textbf{SDXL} & \textbf{SDXL-L} & \textbf{SDXL-T} & \textbf{LCM} & \textbf{PixArt} & \textbf{Cascade} & \textbf{PG2.5} \\
        \midrule
        \textbf{Occupational} & 0.877154 & 0.867155 & 0.867045 & 0.8706605 & 0.736396 & 0.853782 & 0.8543408 \\
        \textbf{Business} & 0.858588 & 0.842781 & 0.849158 & 0.8744108 & 0.804484 & 0.840426 & 0.821007 \\
        \textbf{Science} & 0.884709 & 0.873604 & 0.869268 & 0.8663098 & 0.873176 & 0.875211 & 0.8679658 \\
        \textbf{Legal} & 0.85423 & 0.854885 & 0.865395 & 0.852033 & 0.74844 & 0.862755 & 0.8232336 \\
        \textbf{Education} & 0.878397 & 0.846817 & 0.851358 & 0.855178 & 0.8060857 & 0.853311 & 0.846761 \\
        \textbf{Sports} & 0.863081 & 0.868973 & 0.844147 & 0.8615025 & 0.8387 & 0.85999 & 0.8402564 \\
        \textbf{Arts} & 0.874567 & 0.843933 & 0.856942 & 0.8594696 & 0.638935 & 0.85038 & 0.8407661 \\
        \textbf{Healthcare} & 0.858845 & 0.844196 & 0.855767 & 0.8631429 & 0.847893 & 0.856434 & 0.8415847 \\
        \textbf{Protective} & 0.879518 & 0.889476 & 0.877201 & 0.8818702 & 0.87632 & 0.884892 & 0.869311 \\
        \textbf{Food} & 0.88156 & 0.876571 & 0.873247 & 0.8643314 & 0.8192 & 0.858391 & 0.8283826 \\
        \textbf{Sales} & 0.857379 & 0.857203 & 0.862794 & 0.8656348 & 0.748625 & 0.827692 & 0.8475222 \\
        \textbf{Construction} & 0.895013 & 0.895732 & 0.888275 & 0.8915785 & 0.889572 & 0.896732 & 0.879111 \\
        \textbf{Production} & 0.851386 & 0.846115 & 0.85872 & 0.8545872 & 0.896371 & 0.899712 & 0.8498274 \\
        \textbf{Transportation} & 0.899491 & 0.882141 & 0.874068 & 0.8848437 & 0.886543 & 0.878832 & 0.8778832 \\
        \textbf{Other} & 0.897983 & 0.864047 & 0.861606 & 0.8650834 & 0.856334 & 0.868112 & 0.8603621 \\
        \textbf{Unofficial} & 0.896853 & 0.911507 & 0.895368 & 0.895972 & 0.874442 & 0.905342 & 0.8993091 \\
        \bottomrule
    \end{tabular}
    }
    \caption{Qualitative results of the implicit bias score for the acquired attribute \textit{occupation} of the category level}
\end{table}

\begin{table}[H]
    \centering
    \resizebox{0.9\textwidth}{!}{ % 调整表格宽度以适应页面宽度
    \begin{tabular}{lccccccc}
        \toprule
        \textbf{Attribute} & \textbf{SDXL} & \textbf{SDXL-L} & \textbf{SDXL-T} & \textbf{LCM} & \textbf{PixArt} & \textbf{Cascade} & \textbf{PG2.5} \\
        \midrule
        \textbf{Model} & 0.761634 & 0.717485 & 0.729362 & 0.7033298 & 0.699121 & 0.75455 & 0.7220617 \\
        \textbf{Gender} & 0.862241 & 0.839266 & 0.837606 & 0.8457786 & 0.839024 & 0.883681 & 0.921333 \\
        \textbf{Race} & 0.684828 & 0.622408 & 0.656234 & 0.6532354 & 0.596046 & 0.699239 & 0.698832 \\
        \textbf{Age} & 0.636066 & 0.653465 & 0.588574 & 0.5452472 & 0.631765 & 0.586044 & 0.5642415 \\
        \bottomrule
    \end{tabular}
    }
    \caption{Qualitative results of the explicit bias score at top level.}
\end{table}
\begin{table}[H]
    \centering
    \resizebox{0.9\textwidth}{!}{ % 调整表格宽度以适应页面宽度
    \begin{tabular}{lccccccc}
        \toprule
        \textbf{Attribute} & \textbf{SDXL} & \textbf{SDXL-L} & \textbf{SDXL-T} & \textbf{LCM} & \textbf{PixArt} & \textbf{Cascade} & \textbf{PG2.5} \\
        \midrule
        \textbf{Char} & 0.748272 & 0.783382 & 0.709279 & 0.722682 & 0.752047 & 0.731806 & 0.695863 \\
        \textbf{Oc} & 0.808074 & 0.741146 & 0.772474 & 0.723561 & 0.736396 & 0.801255 & 0.752365 \\
        \textbf{SR} & 0.570398 & 0.545327 & 0.575784 & 0.580886 & 0.481180 & 0.571793 & 0.586911 \\
        \bottomrule
    \end{tabular}
    }
    \caption{Qualitative results of the explicit bias score at acquired attribute level.}
\end{table}
\begin{table}[H]
    \centering
    \resizebox{0.9\textwidth}{!}{ % 调整表格宽度以适应页面宽度
    \begin{tabular}{lccccccc}
        \toprule
        \textbf{Attribute} & \textbf{SDXL} & \textbf{SDXL-L} & \textbf{SDXL-T} & \textbf{LCM} & \textbf{PixArt} & \textbf{Cascade} & \textbf{PG2.5} \\
        \midrule
        \textbf{Char} & 0.748272 & 0.783382 & 0.709279 & 0.7226816 & 0.752047 & 0.731806 & 0.6958632 \\
        \textbf{Positive} & 0.747336 & 0.791424 & 0.715422 & 0.7287691 & 0.753165 & 0.726629 & 0.7025604 \\
        \textbf{Negative} & 0.74982 & 0.770062 & 0.709433 & 0.7125993 & 0.750195 & 0.74038 & 0.68477 \\
        \bottomrule
    \end{tabular}
    }
    \caption{Qualitative results of the explicit bias score for the acquired attribute \textit{characteristic}.}
\end{table}
\begin{table}[H]
    \centering
    \resizebox{0.9\textwidth}{!}{ % 调整表格宽度以适应页面宽度
    \begin{tabular}{lccccccc}
        \toprule
        \textbf{Attribute} & \textbf{SDXL} & \textbf{SDXL-L} & \textbf{SDXL-T} & \textbf{LCM} & \textbf{PixArt} & \textbf{Cascade} & \textbf{PG2.5} \\
        \midrule
        \textbf{Oc} & 0.808074 & 0.741146 & 0.772474 & 0.7235605 & 0.736396 & 0.801255 & 0.7523649 \\
        \textbf{Business} & 0.83744 & 0.751674 & 0.813462 & 0.7399 & 0.78165 & 0.827864 & 0.804281 \\
        \textbf{Science} & 0.79611 & 0.707515 & 0.768942 & 0.6506 & 0.728948 & 0.803944 & 0.733186 \\
        \textbf{Legal} & 0.8321 & 0.765572 & 0.780958 & 0.782395 & 0.745676 & 0.836009 & 0.77579 \\
        \textbf{Education} & 0.847111 & 0.757979 & 0.80826 & 0.7530484 & 0.75338 & 0.832759 & 0.780054 \\
        \textbf{Sports} & 0.774061 & 0.706943 & 0.740318 & 0.6718157 & 0.6936 & 0.803702 & 0.709054 \\
        \textbf{Arts} & 0.84751 & 0.81539 & 0.79991 & 0.780904 & 0.749536 & 0.794052 & 0.76609 \\
        \textbf{Healthcare} & 0.774139 & 0.673 & 0.772538 & 0.717354 & 0.715774 & 0.826331 & 0.730767 \\
        \textbf{Protective} & 0.71238 & 0.683518 & 0.694483 & 0.666653 & 0.664653 & 0.755274 & 0.67327 \\
        \textbf{Food} & 0.801697 & 0.688208 & 0.724755 & 0.69841 & 0.738618 & 0.81335 & 0.747214 \\
        \textbf{Sales} & 0.8052 & 0.72743 & 0.792487 & 0.7238 & 0.762216 & 0.822813 & 0.789124 \\
        \textbf{Construction} & 0.770406 & 0.722921 & 0.711622 & 0.6657627 & 0.724158 & 0.7759 & 0.7074 \\
        \textbf{Production} & 0.773377 & 0.7202 & 0.75813 & 0.673737 & 0.7325 & 0.782291 & 0.7122 \\
        \textbf{Transportation} & 0.810363 & 0.752469 & 0.765053 & 0.7289847 & 0.72987 & 0.786046 & 0.749405 \\
        \textbf{Other} & 0.834222 & 0.791799 & 0.817968 & 0.763387 & 0.74717 & 0.821354 & 0.762697 \\
        \textbf{Unofficial} & 0.762303 & 0.753113 & 0.7194 & 0.7007235 & 0.716219 & 0.726668 & 0.71144 \\
        \bottomrule
    \end{tabular}
    }
    \caption{Qualitative results of the explicit bias score for the acquired attribute \textit{occupation}.}
\end{table}
\begin{table}[H]
    \centering
    \resizebox{0.9\textwidth}{!}{ % 调整表格宽度以适应页面宽度
    \begin{tabular}{lccccccc}
        \toprule
        \textbf{Attribute} & \textbf{SDXL} & \textbf{SDXL-L} & \textbf{SDXL-T} & \textbf{LCM} & \textbf{PixArt} & \textbf{Cascade} & \textbf{PG2.5} \\
        \midrule
        \textbf{SR} & 0.570398 & 0.545327 & 0.575784 & 0.5808856 & 0.48118 & 0.571793 & 0.5869107 \\
        \textbf{Intimate} & 0.475164 & 0.45316 & 0.48225 & 0.5709764 & 0.342996 & 0.449526 & 0.587393 \\
        \textbf{Instructional} & 0.723089 & 0.65004 & 0.710909 & 0.6841308 & 0.659111 & 0.70602 & 0.692179 \\
        \textbf{Hierarchical} & 0.69611 & 0.648119 & 0.682332 & 0.6816366 & 0.594869 & 0.687607 & 0.6907154 \\
        \bottomrule
    \end{tabular}
    }
    \caption{Qualitative results of the explicit bias score for the acquired attribute \textit{social relation}.}
\end{table}
\begin{table}[H]
    \centering
    \resizebox{0.9\textwidth}{!}{ % 调整表格宽度以适应页面宽度
    \begin{tabular}{lccccccc}
        \toprule
        \textbf{Attribute} & \textbf{SDXL} & \textbf{SDXL-L} & \textbf{SDXL-T} & \textbf{LCM} & \textbf{PixArt} & \textbf{Cascade} & \textbf{PG2.5} \\
        \midrule
        \textbf{Model} & 0.612138209 & 0.620129205 & 0.569934852 & 0.624898235 & 0.616241859 & 0.617573949 & 0.630034 \\
        \textbf{Gender} & 0.62467837 & 0.578166303 & 0.545608051 & 0.60142146 & 0.632191062 & 0.629328861 & 0.664900445 \\
        \textbf{Race} & 0.653047182 & 0.721388385 & 0.621537737 & 0.70951493 & 0.671283931 & 0.666262291 & 0.661665 \\
        \textbf{Age} & 0.558689074 & 0.560832927 & 0.542658768 & 0.5637583 & 0.545250583 & 0.557130694 & 0.56353677 \\
        \bottomrule
    \end{tabular}
    }
    \caption{Qualitative results of the manifestation factor \(\eta\).}
\end{table}
\section{Ethical Explanation}
\label{Ethical}
As our research involves human evaluators and the benchmark is designed for evaluating bias, ethical considerations are crucial.\\
All evaluators were fully informed about the purpose of our study and potential offensive content including gender, race and age discrimination. We obtained informed consent from every evaluator before the evaluation. Evaluators were anonymized in any reports or publications resulting from the study, ensuring the personal information security. Evaluators received comprehensive training on how to perform evaluations effectively and ethically. The design of the datasheet for evaluation was inspired by the guideline by \cite{gebru2021datasheets}. A template of the datasheet for human evaluation is also provided in our repository.\\
For the ethical impacts of our work, we consider how FAIntbench might influence future practices in the bias evaluation of T2I models. We believe that transparency in the evaluation process and datasets is crucial, influenced by \cite{larsson2020transparency}. Therefore, we decide to open-source FAIntbench, including the dataset and evaluation metrics, under the GPL license. Our commitment extends to maintaining transparency in how the evaluation results are utilized, with the aim of encouraging open discussions in the bias evaluation of T2I models.

\section{Limitations and Potential Improvements to FAIntbench}
\label{Limitation}
Based on the initial results, we propose the following potential future directions to delve deeper into the biases and their impacts within large-scale T2I models:
\\
\textbf{Test More Models}: Due to budget constraints,  models like DALL-E V2/V3 (\cite{ramesh2022hierarchical:8,betker2023improving}), Midjourney v6, and Stable Diffusion V3 (\cite{esser2024scaling:9}) have not been tested. Like all existing bias benchmarks, FAIntbench dosen't evaluate models that are specifically optimized for debiasing by recent research (\cite{clemmer2024precisedebias,gandikota2024unified,schramowski2023safe,lyu2023attention}). Future studies will aim to include more models, especially those tailored to decrease biases.
\\
\textbf{Develop a More Comprehensive Algorithm}: Our preliminary findings utilize only results from implicit prompts to calculate the manifestation factor. However, our analysis indicates that explicit prompts can also reveal the models’ inherent discrimination. Developing an optimized algorithm can lead to a more accurate manifestation factor.
\\
\textbf{Develop a User-Friendly Interface}: We have provided a set of adjustable prompts and evaluation metrics for researchers, which enhance the utility of FAIntbench. Currently, we offer guidelines for modifying datasets and code, which can be somewhat complex. In the future, we plan to create a more integrated modification tool that allows researchers to update FAIntbench by simply inputting a JSON file with the changes, thus reducing their workload and facilitating broader adoption of FAIntbench.
\\
\textbf{Solve the Increasing Biases from Distillation}: We have conducted preliminary studies and discussions on the side effects of distillation on bias using FAIntbench. Given the increasing focus on distillation techniques in the text-to-image domain (\cite{song2024sdxs,lin2024sdxl,luo2023lcm}), we believe that further research and tailored solutions for the increasing biases are crucial.

\end{document}